\RequirePackage{fix-cm}
\documentclass[smallextended]{svjour3}       
\smartqed  
\usepackage{graphicx}
\usepackage{multirow}
\usepackage{subfigure}
\usepackage{amsmath}
\usepackage{booktabs}

\usepackage{algorithm}  
\usepackage{algorithmicx}  
\usepackage{algpseudocode}  
\usepackage{amsmath} 
\usepackage{amssymb}
\floatname{algorithm}{Algorithm}

\begin{document}

\title{RLINK: Deep Reinforcement Learning for User Identity Linkage
}


\author{Xiaoxue Li         \and Yanan Cao           \and Yanmin Shang     \and
        Yangxi Li          \and
        Yanbing Liu        \and
        Jianlong Tan
}


\institute{Xiaoxue Li \at
	          College of Cyberspace Security, University of Chinese Academy of Sciences, Beijing, China\\
	          Institute of Information Engineering，Chinese Academy of Sciences, Beijing, China\\
              \email{lixiaoxue@iie.ac.cn}           
           \and
           Yanan Cao \at
           Institute of Information Engineering，Chinese Academy of Sciences, Beijing, China\\
           \email{caoyanan@iie.ac.cn}
           \and
           Yanmin Shang \at
           Institute of Information Engineering，Chinese Academy of Sciences, Beijing, China\\
           \email{shangyanmin@iie.ac.cn}            
           \and
           Yangxi Li \at
           National Computer network Emergency Response technical Team, Beijing, China\\ 
           \email{liyangxi@outlook.com}
           \and
           Yanbing Liu \at
           Institute of Information Engineering，Chinese Academy of Sciences, Beijing, China\\
           \email{liuyanbing@iie.ac.cn}
           \and
           Jianlong Tan\at
           Institute of Information Engineering，Chinese Academy of Sciences, Beijing, China\\
           \email{tanjianlong@iie.ac.cn}                             
}

\date{Received: date / Accepted: date}

\maketitle

\begin{abstract}
User identity linkage is a task of recognizing the identities of the same user across different social networks (SN). Previous works tackle this problem via estimating the pairwise similarity between identities from different SN, predicting the label of identity pairs or selecting the most relevant identity pair based on the similarity scores. However, most of these methods ignore the results of previously matched identities, which could contribute to the linkage in following matching steps. To address this problem, we convert user identity linkage into a sequence decision problem and propose a reinforcement learning model to optimize the linkage strategy from the global perspective. Our method makes full use of both the social network structure and the history matched identities, and explores the long-term influence of current matching on subsequent decisions. We conduct experiments on different types of datasets, the results show that our method achieves better performance than other state-of-the-art methods. 
\end{abstract}

\section{Introduction}
\label{intro}

User Identity Linkage (UIL), which aims to recognize the identities (accounts) of the same user across different social platforms, is a challenging task in social network analysis. Nowadays, many users participate in multiple on-line social networks to enjoy more services. For example, a user may use Twitter and Facebook at the same time. However, on different social network platforms, the same user may register different accounts, have different social links and deliver different comments. If different social networks could be integrated together, we could create an integrated profile for each user and achieve better performance in many practical applications, such as link prediction and cross-domain recommendation. So, UIL has recently received increasing attention both in academia and industry. 

Most of previous works consider UIL as a one-to-one alignment problem \cite{one,IONE,PALE,cosnet}, i.e., each user has at most one identity in each social network. Matching models \cite{34,53}, label propagation algorithms \cite{23,69,32,48,40,78} and ranking algorithms \cite{IONE,MAH,PALE,deeplink} are commonly used to address this task.
These methods generally calculate pairwise similarity between identities and select the most relevant identity pairs according to the similarity score. In more recent works, identity similarity is computed based on user embedding \cite{PALE}, which encodes the main structure of social networks or other features into a low-dimensional and density vector. In most of these methods, each identity pair is matched independently, i.e., one predicted linkage of the identity pair would not be effected by any other.

\begin{figure}[ht]
	\centering  
	\includegraphics[width=8cm]{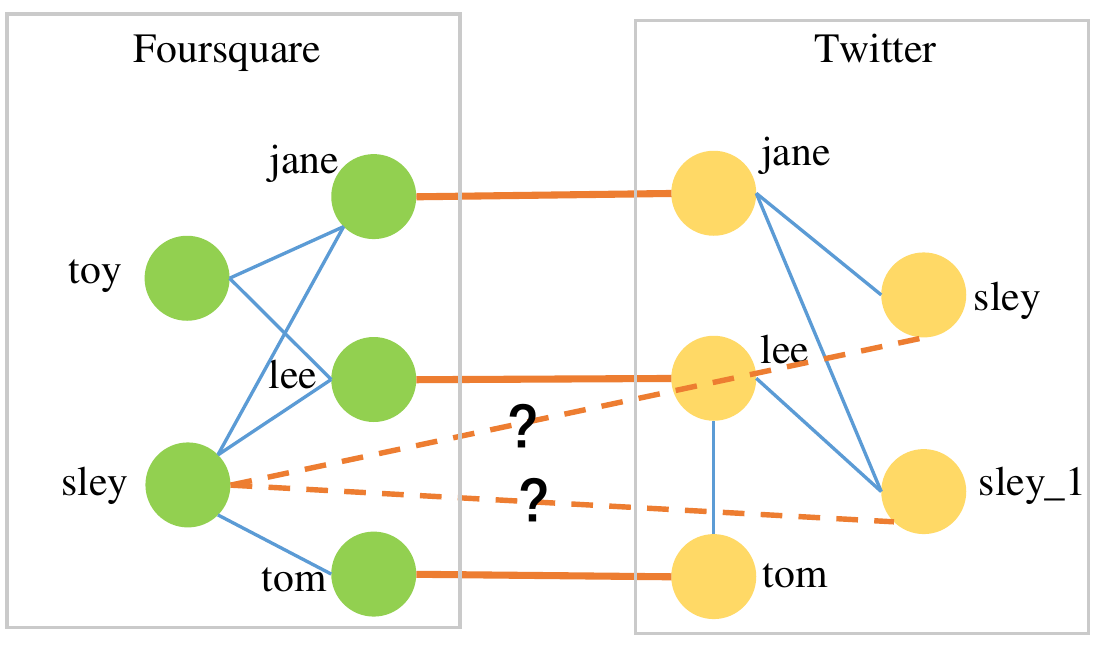}  
	\caption{An example of User Identity Linkage. The blue link represents friend relation in social network, and orange line represents matched identity pair. User identity sley (in Foursquare) has two candidate identities in Twitter: sley and sley\_1. If considering previously matched information, sley in Foursquare is similar to sley\_1 in Twitter because they share more similarity friends.}  
	\vspace{-1em}
	\label{example}  
\end{figure}

However, we have an intuition that, the predicted linkages are inter-dependent and the previously matching would have a long-term influence on the subsequent one. This influence is two-fold: (i) if the preceding decisions are right, they would bring in positive auxiliary information to guide the following linkage. For example, if two user from two different social networks share the same friends (their friends have been matched), they are more likely to be matched in the subsequent linkage, as shown in Figure \ref{example}; (ii) the previously matched user identity could not be chosen in the subsequent matching process according to the \textit{one-to-one constraint} \cite{one}, 
Although some previous label propagation based works have made primary attempts on using this influence, they just gave the fixed greedy strategy to iteratively select the candidate identity pair \cite{32,48,40,78}. Neither of them re-calculate the pairwise similarity or dynamically adjust the linkage strategy after each matching step due to the high complexity.

\begin{figure}[t]
	\centering  
	\includegraphics[width=\textwidth]{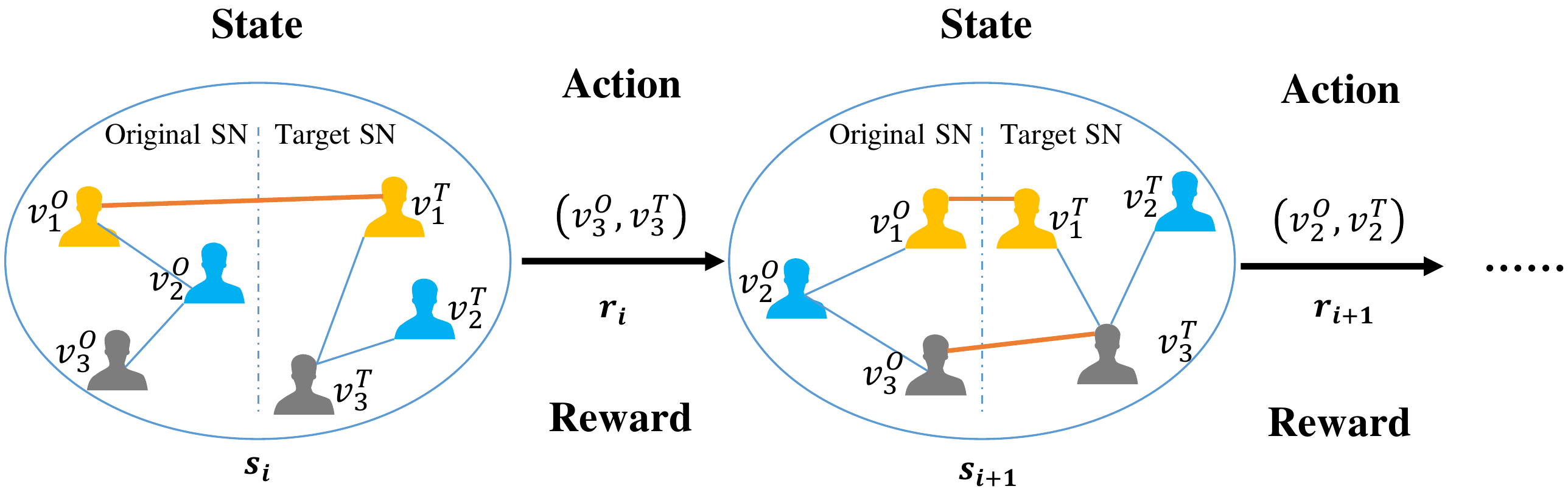}  
	\caption{A Procedure of Reinforcement Learning based User Identity Linkage. The blue link represents friend relation in social network, and orange line at $S_i$ represents matched identity pair. At time $i$, agent generates a pair of matching identities as action according to current state. After environment performs this action, the state would be changed at time $i+1$ and next action can be generated based on $S_{i+1}$.}  
	\vspace{-1em}
	\label{procedure}  
\end{figure}

In order to model this long-term influence effectively, we novelly consider UIL as a Markov Decision Process and propose a deep reinforcement learning (RL) framework RLink to automatically match identities in two different social networks. Figure \ref{procedure} illustrates the overall RLink process.  In each time stamp, the \textbf{state} consists of two social network structures and previously matched identity pairs. According to the current state, the agent perform an \textbf{action}, i.e., generating an identity pair from the candidates. After performing the action,  the \textbf{state} would be changed at the next time and a \textbf{reward} would be fed back to the agent to adjust its policy. Because the action space is large and dynamic in the UIL process, we adapt an Actor-Critic framework, in which the Actor network generates a deterministic action based on current state and the Critic network evaluates the quality of this action-state pair. Concretely, for each state, the Actor firstly encodes the network structure and history decisions to a latent vector representation, and then decodes this vector into an action in the identity embedding space. Based on this, our model could generate the matching sequence automatically.

Deep RL model, which learns to directly optimize the overall evaluation metrics, works much better than models which learn with loss functions that just evaluate a particular single decision \cite{Sigma,cosnet}. However, there has been spots of attempts on applying RL in the social network analysis field \cite{graph,dynamic,Peyravi}. To the best of our knowledge, we are the first to design a deep RL model for user identity linkage. And in this paper, our RL model is able to produce more accurate results by exploring the long-term influence of independent decisions.

The contributions of this paper can be summarized as follows:
\begin{itemize} 
	\item  We are the first to consider UIL as a sequence decision problem and innovatively propose a deep reinforcement learning based model in this task, which generates the matching sequence automatically. 
	\item The proposed model makes full use of the previous matched identity pairs which may impact on the subsequent linkage, and makes decisions from a global perspective. 
	\item We conduct extensive experiments on three pairs of real-word datasets to show that our method achieves better performance than other state-of-the-art solutions to the problem. 
\end{itemize}

\section{The Proposed Model}
\subsection{Preliminary}
\label{sec:1}
Let $\mathcal{G}=(\mathcal{V},\mathcal{E})$ represents an online social network, where $\mathcal{V}=\{v_1,v_2,...,v_N\}$ is the set of user identities and $\mathcal{E}\in \mathcal{V}\times \mathcal{V}$ is the set of links in the network. Given two online social network $\mathcal{G}^O$ (original network) and $\mathcal{G}^T$ (target network), the task of user identity linkage is to identify hidden user identities pairs across $\mathcal{G}^O$ and $\mathcal{G}^T$. Here, we have a set of node pair between $\mathcal{G^O}$ and $\mathcal{G}^T$ to represent given alignment information (ground-truth), denoted as $\mathcal{B}$ 

In this work, we consider UIL task as a markov decision process in which the linkage agent interacts with environment over a sequence of steps. We use $s \in \mathcal{S}$ to denote the current state, which consists of the network structure and matched identity pairs. Action, which is denoted as $a \in \mathcal{A}$, is a pair of identities. The action space of UIL is comprised of all potential identity pairs, the size of which is $|\mathcal{V}^O|\times |\mathcal{V}^T|$. At each step, the agent generates an action $a$ based on $s$, and would receive a reward $r(s,a)$ according to the given alignment information. The goal of RL is to find a linkage policy $\pi : \mathcal{S} \leftarrow \mathcal{A}$, which can maximize the cumulative reward for linkage.

\begin{figure*}[h]
	\centering  
	\includegraphics[width=1\linewidth]{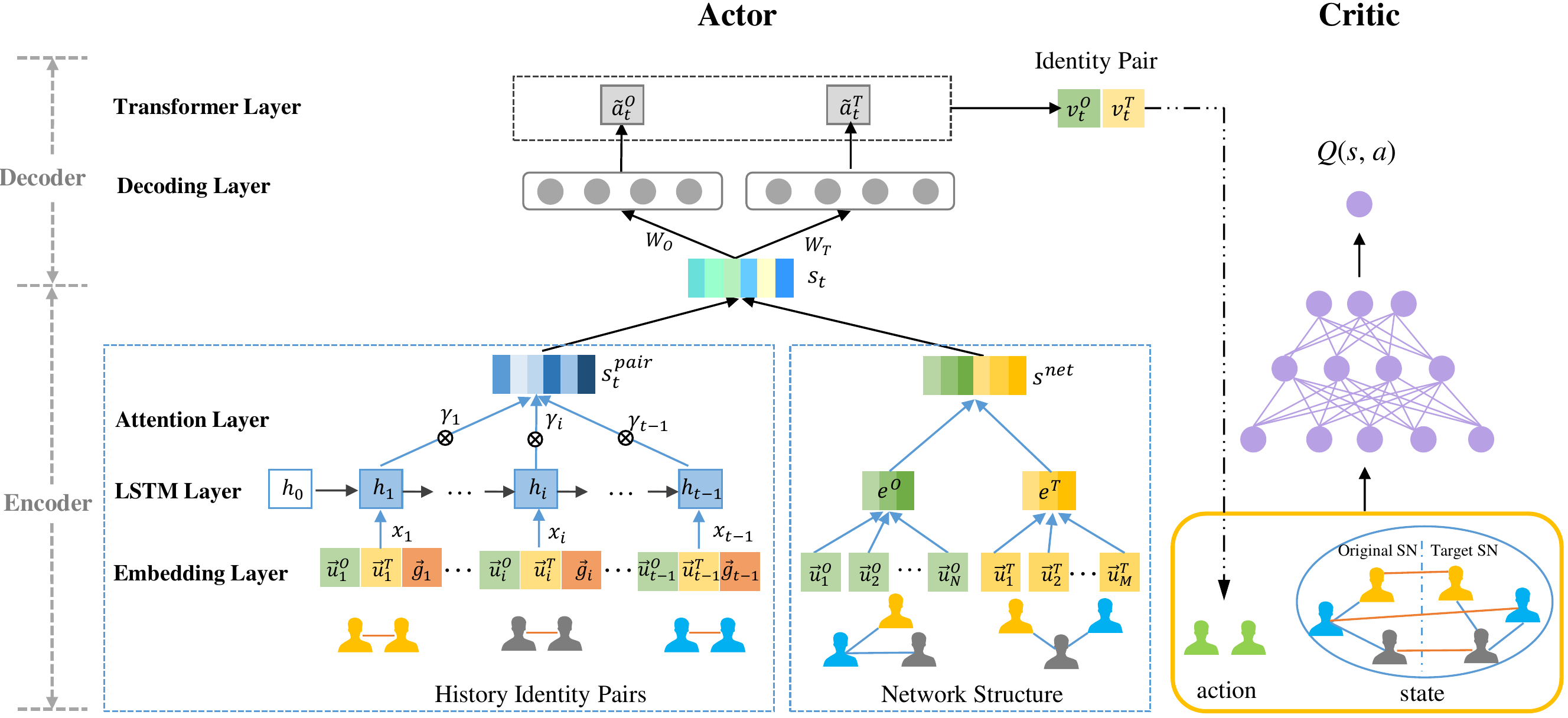}  
	\caption{The framework of Actor-Critic network, where the actor is comprised of an Encoder-Decoder architecture and the critic is DQN. The inputs of Actor are history identity pairs and network structure, where history identity pairs were generated by our Actor-Critic network before current epoch (See Eq.(4)). Remarkably, $h_0$ is a zero vector. Then this action and current state are input onto the Critic to evaluate the quality of this action.    
	}  
	\label{framework}   
\end{figure*}

Due to the large and dynamic action space, the policy-based
RL \cite{fangzheng,policy} which compute probability distribution of
every action and the Q-learning \cite{Atari,hybrid} which evaluate the
value of each potential action-state pair are time-consuming. To reduce the computational cost, we adapt the Actor-Critic framework, where the actor inputs the current state $s$ and aims to output a deterministic action, while the critic inputs only this state-action pair rather than all potential state-action pairs. The architecture of the Actor-Critic network is shown in Figure \ref{framework}.

\subsection{Architecture of Actor Network}
The goal of our Actor network is to generate an action (one matching identity pair) according to the current state. We propose an Encoder-Decoder architecture to achieve this goal.

\subsubsection{Encoder for Current State}
Encoder aims to generate the representation of current state, which contains two types of information: network structure and previous matched identity pairs. In order to integrate these information, we apply two encoding mechanisms to respectively encode the network structure and the matched pairs, as shown in Figure \ref{framework}. 

In our model, each identity $v_k$ ($k\in (1,N)$)is represented as a low dimensional and dense vector, which is denoted as $\vec u_k$ and $\vec u_k \in R^{d}$ ($d$ is the dimension of the identity embedding). These identity embeddings are pre-trained by Node2vec \cite{Node2vec}. To represent the social network structure, we weighted sum all identity embedding to generate the network embedding $e$, which is inspired by \cite{embedding}.  
\begin{equation}
e = \sum_{v_k\in \mathcal{V}} \alpha_k \vec u_k,
\end{equation}
where $\alpha_k$ denotes the weight of the identity $v_k$. Here, we define $\alpha_k$ as $\frac{\zeta}{\zeta+ d(v_k)}$ (similar to \cite{embedding}), where $\zeta$ is a constant and $ d(v_k)$ represents the degree of identity $v_k$. We denote the representation of original network and target network as $e^O$ and $e^T$, then we concatenate them to get $s^{net}$ and $s^{net} \in R^{2d}$. That is:
\begin{displaymath}
s^{net} = concat(e^O, e^T).
\end{displaymath}

At time $t$, we need to encode all previously matched identity pairs from time 1 to $t-1$, i.e.,  an action sequence $\{a_1$,$a_2$, ..., $a_{t-1}\}$.  In each action $a_i$ $(i \in [1, t-1])$, the identity from $\mathcal{G}^O$, $\mathcal{G}^T$ are respectively denoted as $v^{O}_{i}$ and $v^{T}_{i}$, and their embedding representation are denoted as $\vec u^{O}_{i}$ and $\vec u^{T}_{i}$. Besides, we take the feedback of $a_i$, which is denoted as $g_{i}$, into account. $g_i$ is a one-hot vector, in which the value of each dimension is equal to the immediate reward at the corresponding time stamp. 
Through a neural network layer, we get the encoding vector of $g_{i}$ as $\vec g_{i}$:
\begin{equation}
\vec g_{i} = \varphi(W_G g_{i} + b_G),
\end{equation} 
where $g_{i}\in {R}^{|g|}$ and $\vec g_{i}\in {R}^{|G|}$, and ${R}^{|g|}$ is equal to the number of steps in each episode and ${R}^{|G|}$ depends on the dimension of $W_G$. 

Then, we get the representation of one history matched pair at time $i$ by concatenating $\vec u^{O}_{i}, \vec u^{T}_{i}$ and $\vec g_{i}$:
\begin{equation}
x_{i} = concat(\vec u^{O}_{i}, \vec u^{T}_{i}, \vec g_{i}).
\end{equation}
where the dimensional of $x_{i}$ is $|G|+2d$. 

In order to capture the long-term influence of the previously matched pairs, we use a Long Short-Term Memory (LSTM) network to encode the history linkage sequence into a fixed-size vector:
\begin{equation}
h_i = LSTM(x_i, h_{i-1}),
\end{equation}
Furthermore, to distinguish different contribution of the previous actions, we employ attention mechanism \cite{attention}, which allows model to adaptively focus on different parts of the input:
{-0.5em}
\begin{equation}
s^{pair}_t = \sum_{i=1}^{t-1}\gamma_i h_i,
\end{equation}
where the dimension of $s_t^{pair}$ is equal to $s^{net}$, and we leverage a location-based attention mechanism\cite{local attention} to compute $\gamma_i$ from the hidden state $h_i$
\begin{displaymath}
\gamma_i = \frac{exp(w_\gamma h_i + b_\gamma)}{\sum_{j}exp(w_\gamma h_j + b_\gamma)}
\end{displaymath}

Then, the embedding of current state can be represented as follows: 
\begin{equation} 
{s}_{t} = S_t^{pair} + s^{net},
\end{equation}  

where ${s}_{t}\in R^{2d}$, $s^{pair}_t$ and $s^{net}$ represent the embedding of history matching information and network structure respectively. 

It is noteworthy that, at time 1, the current state just contains two social network structures because there is not matched identity pair. That is to say, $h_0 = 0$, $s^{pair}_t$ is equal to zero vector and $s_{t}$ is equal to $s^{net}$. 

Note that, the decoded $\widetilde a^{O}_{t}$ and $\widetilde a^{T}_{t}$ maybe not in the identity embedding space. Thus we need to map them into the real embedding space via the transformation \cite{Deeppage}. As mentioned above, the action space is very large. In order to correctly map $\widetilde {a}_{t}$ into the validate identity, we select the most similar $\vec u_{t} \in U$ as the valid identity-embedding. In this work, we compute the cosine similarity to get the valid identity in given networks: 
\begin{equation}
v^{O}_{t} = \max_{v^{O} \in \mathcal{V}^O}  \left(\vec u^{O}_{t}\right)^\mathrm{T} \cdot \frac{U^O}{\lVert U^O  \rVert},
\end{equation}
\begin{equation}
v^{T}_{t} = \max_{v^{T} \in \mathcal{V}^T}  \left(\vec u^{T}_{t}\right)^\mathrm{T} \cdot \frac{U^T}{\lVert U^T  \rVert},
\end{equation}
where $\{{v}_{t}^O, {v}_{t}^T\}$ represents the valid identity pair $a_t$. 
We pre-compute the value of $\frac{U^O}{\lVert U^O  \rVert} $ and $\frac{U^T}{\lVert U^T  \rVert}$ to decrease the computational cost. 
Since the right alignment identities could be chosen in the following steps, we ignore those identities if they are correctly matched in the previous steps.

\subsubsection{Reward}
The agent generates the new identity pair and receives 
the immediate reward $r_{tm}$ from networks, which is also the feedback of this identity pair,
\begin{displaymath}
r_{tm} = \begin{cases}
1,& a_i \in Groundtruth; \\
-1,& \text{else},
\end{cases}
\end{displaymath}
where $Groundtruth$ is a set of known aligned identity pairs. Since current action result has a long-term impact on subsequent decisions, we introduce a discount factor $\lambda$ to metric the weight of reward:
\begin{equation}
r_t = \lambda_t r_{tm},
\end{equation}
where $\lambda_t \in \{0,1\}$ represents how much influence the action $a_t$ will generate on the following steps. Simply, $\lambda_t$ is defined as $\frac{1}{\boldsymbol{t}}$, where $t$ represents the current time stamp.    

\subsection{Architecture of Critic Network}
Critic Network aims to judge whether the action ${a_{t}}$ generated by Actor suits the current state ${s_{t}}$. Generally, the Critic is designed to learn an action-value function $Q(s,a)$, while the actor updates its' parameters in a direction of improving performance to generate next action according to $Q(s,a)$ in the following steps. However, in real UIL, the state and action space is enormous and many state-action pairs may not appear in the real traces, which makes it hard to update their Q-values. Thus, we choose Deep Neural Network as an approximator to estimate the action-value function. In this work, we refer to a neural network as Deep Q-value function (DQN)\cite{DQN}.

Firstly, we need to feed user's current state $s$ and action $a$ into the DQN. To generate user's current state $s$, the agent follows the same strategy from Eq.(1) to Eq.(6). As for action $a$, we utilize the same strategy in decoder to compute a low-dimensional dense action vector $a$. Then, this action is evaluate by the DQN, which returns the Q-value of this state-action pair $Q(s, a)$.

\subsection{Training and Test}
Generally, we utilize DDPG\cite{DDPG} algorithm to train the proposed Actor-Critic framework, which has four neural network: critic, actor, critic-target and actor-target, where target networks are the copy of critic and actor respectively. The critic is trained by minimizing the mean squared error loss with the corresponding target given by:
\begin{equation}
L(\beta_c) = E_{s, a, R, s'}[(R+\rho Q_{\beta_c'}(s',f_{\theta_{\pi^{\prime}}}(s')) -Q_{\beta_c}(s, 
a))^2],
\end{equation} 
where $\rho$ is the discount factor in RL, $R$ is the accumulative reward and $s'$ is the previous state, $f_{\theta_{\pi^{\prime}}}(s') = a'$. Besides, $\beta_c$ and $\theta_{\pi}$ represent all parameters in Critic and Actor respectively, and $f_{\theta_{\pi}}$ represents the policy. The Critic is trained from samples stored in a replay buffer\cite{replay}. Similarity, actions also stored in the replay buffer generated by the strategy in Actor decoder section. This allows the learning algorithm to dynamic leverage the information of which action was actually executed to train the critic.

The first term $R+\rho Q_{\beta_c'}(s',f_{\theta_{\pi^{\prime}}}(s'))$ in Eq.(9) is the output of target, namely $y$, for current iteration. And parameters from the previous iteration $ \theta_{\pi'}$ are fixed when optimizing the loss function $L(\beta_c)$. Computing the full expectations' gradient are not efficient. Thus, we optimize the loss function by Stochastic Gradient Descent (SGD). The derivatives of loss function $L(\beta_c)$ with respective to parameters $\beta_{c}$ are presented as follows:
\begin{equation}
	\triangledown_{\beta_{c}}L(\beta_{c})= E_{s,a,r,s'}[R+\rho Q_{\beta_c}(s',f_{\theta_{\pi^{\prime}}}(s'))-Q_{\beta_c}(s, 
	a)\triangledown_{\beta_{c}}Q_{\beta_c}(s, 
	a)]
\end{equation}

The Actor is updated by using the policy gradient:
\begin{equation}
	\triangledown_{\theta_{\pi}} \approx E_s[\triangledown_a Q_{\beta_c}(s, 
	a)\triangledown_{\theta_{\pi}f_{\theta_{\pi}}}]
\end{equation}

The training algorithm for the proposed framework RLink is presented in Algorithm 1. In each iteration, there are two stages, i.e., 1) generating an action (lines 8-11), and 2) parameter updating (lines 13-17). For generating an action, given the current state $s_t$, the agent firstly encode current state as a vector (line 8) and then generate a pair of nodes according to this vector (line 9); then the agent observes the reward $r_t$ and update state to $s_{t+1}$ (line 10); finally the agent stores transitions $(s_t,a_t,r_t,s_(t+1)) $ into replay buffer $\mathcal{D}$. For the parameter updating stage: the agent samples mini-batch of transitions $(s,a,r,s')$ from $\mathcal{D}$ (line 13), and then updates parameters of Actor and Critic following a standard DDPG procedure (lines 14-16).
Finally, the parameters of target network $\beta_{c'}$ and $\theta_{\pi'}$ is updated via the soft update way (line 17).

To evaluate the performance of our model, the test procedure is designed as an online test method, which is similar to the action generation stage in the training procedure. After the training procedure, proposed framework RLink learns parameters $\theta_{\pi}$ and $\beta_{c}$. In each iteration, the agent generates a pair of identity $a_t$ following the trained policy $f_{\theta_{\pi}}$. And then the agent receives the reward $r_t$ from networks and updates the state to $s_{t+1}$.

\begin{algorithm} 
	\caption{ The RLink algorithm.}  
	\label{alg:Framwork}  
	\begin{algorithmic}[1]   
		\State Initial Actor network $f_{\theta_{\pi}}$ and critic network $Q_{\beta_{c}}$ with random weights;
		\State Initial target network $f_{\theta_{\pi^{\prime}}}$ and $Q_{\beta_{c'}}$ with weights $\theta_{\pi^{\prime}}\leftarrow\theta_{\pi}$, $\beta_{c'}\leftarrow\beta_{c}$  
		\label{code:fram:feature}  
		\State Initial the capacity of replay buffer $\mathcal{D}$
		\label{code:fram:trainbase}
		\For {$session=1,\Gamma$} 
		\State Receive initial observation state $s_1$  
		\label{code:fram:add}
		\For{$t=1, T$}
		\State \textbf{Stage 1: Generating an action}
		\State Encode current state $s_t$ according to Eq.(1) to Eq.(6)
		\State Generate the valid identity pair according Eq.(7) and Eq.(8) as action $a_t$
		\State Observe the reward $r_t$ according to Eq.(9) and new state $s_{t+1}$ 
		\State Store transition $(s_t,a_t,r_t,s_{t+1})$ in $\mathcal{D}$
		\State \textbf{Stage 2: Parameter updating}
		\State Sample mini-batch $\mathcal{N}$ transitions $(s,a,r,s')$ in $\mathcal{D}$
		\State Set $R+\rho Q_{\beta_c}(s',f_{\theta_{\pi^{\prime}}}(s'))$
		\State Update Critic by minimizing $\frac{1}{\mathcal{N}}\sum_n(y-Q_{\beta_c}(s,a))^2$ according to Eq.(11)
		\State Update Actor using the sampled policy gradient according to Eq.(12)
		\State Update the target network:
				      $\theta_{\pi^{\prime}}\leftarrow\tau\theta_{\pi}+(1-\tau)\theta_{\pi^{\prime}}$;
				      $\beta_{c'}\leftarrow\tau\beta_{c}+(1-\tau)\beta_{c^{\prime}}$ 
		
		\EndFor
		\EndFor	  
		\label{code:fram:classify}   
	\end{algorithmic}  
\end{algorithm}

\section{Experiment}
In this section, we compare our RLink with the state-of-the-art methods on types of cross-network datasets. 

\subsection{Experiment Setup}
\subsubsection{Datasets}

In order to verify our method in different types of networks, we conduct experiments on the following network collections: Foursquare-Twitter, Last.fm-Myspace and Aminer-Linkedin, which are commonly used in the UIL task. The first dataset is provided by \cite{PCT} and others are are collected by\cite{cosnet}. These datasets are introduced as follows and the statistic information is shown in Table \ref{data} which only contain social links as user identities' feature. 

\begin{itemize}
	\item Foursquare-Twitter is a pair of social networks, where users share their current location and other information with others. 
	\item Last.fm-Myspace is a pair of online social networks, where users could search music and share their interested music with others. 
	\item Aminer-Linkedin is a pair of citation networks where contains users' academic achievements and users could search interested community.
\end{itemize}

\begin{table*}
	\caption{Statistics of experimental datasets}
	\label{tab:freq}
	\begin{tabular}{cccl}
		\toprule
		Dataset& User Identities & Social links & Ground truth \\
		\midrule
		Foursquare & 5,120 & 76,972&  \multirow {2}{*}{1,609}      \\
		Twitter & 5,313& 164,920&      \\\midrule
		Last.fm & 136,420 & 1,685,524&  \multirow {2}{*}{1,381}      \\
		Myspace & 854,498& 6,489,736&      \\\midrule
		Aminer &1,053,188 & 3,916,907&\multirow{2}{*}{4,153}  \\
		Linkedin & 2,985,414 & 25,965,384&\\
		\bottomrule
	\end{tabular}
	\label{data}
\end{table*} 

\subsubsection{Comparative Methods}
In this section, to evaluate the performance of RLink for user identity linkage, we choose the following state-of-the-art methods as competitions, including:
\begin{itemize}
	\item IONE \cite{IONE} IONE predicts anchor links by learning the followership embedding and followeeship embedding of a user simultaneously.
	\item DeepLink \cite{deeplink}  DeepLink employs unbiased random walk to generate embeddings, and then use MLP to map users. 
	\item MAG \cite{MAH} MAG uses manifold alignment on graph to map users across networks. 
	\item PAAE \cite{PAAE} PAAE employs an adversarial regularization to capture the robust embedding vectors and maps anchor users with an alignment auto-encoders.
	\item SiGMa \cite{Sigma} SiGMa was designed to align two given network by propagating the confidence score in the matching network. We use the name matching method to generate the seed set and utilize the output scores of SVM as the pairwise similarity. Note that SiGMa is an unsupervised method.
	\item SDM: This method is a simpler deep model, which is similar to typical sequence prediction methods \cite{sequence,BiLSTM} and utilizes LSTM to process sequence matching. The matching sequence is generated via ranking the similarity of embedding. At each step $i$, LSTM predicts an user identity in Target SN based on current user identity embedding and previous hidden information ($h_{i-1}$). Finally, the L2 loss would guide the training and testing process of this model.          	
\end{itemize} 

\subsubsection{Evaluation Metrics}
To perform the user identity linkage, we utilize two standard metrics \cite{review} to evaluate the performance, including \textit{Precision@k (P@k)}, recall and \textit{MAP}. Note that higher the value of each these measures, the better the performance.

\textit{Precision@k} is the metric for evaluating the linking accuracy, defined as:
\begin{equation}
	P@k = \sum_{i}^{n} \mathbf{1}_i\{\textit{success@k}\}/n,
\end{equation}
where $\mathbf{1}_i\{\textit{success@k}\}$ measures whether the positive matching identity exists in $top-k (k<=n)$ list, and $n$ is the number of testing anchor nodes.

The recall is the fraction of the number of real corresponding user pairs that have been found over the total amount of real matched user pairs (Ground-truth $\mathcal{B}$).

\begin{equation}
	Recall = \frac{\textit{success matched pairs}}{\textit{Real user pairs in }\mathcal{B}}
\end{equation}

\textit{MAP} is used for evaluating the ranking performance of the algorithms, defined as:
\begin{equation}
	MAP = (\sum^n \frac{1}{ra})/n,
\end{equation}    
where $ra$ is the rank of the positive matching identity and $n$ is the number of testing anchor nodes.

\subsubsection{Hyper-parameter setting}
For each pair of networks, we first resort the ground truth data set by identity number and then use the first $r$ as the training data and the later $1-r$ as the testing data, where $r$ means the training ratio. The dimension of the identity embedding is set to 128, whether the input of the encoder or the output of the decoder, i.e., $|U| = |G| = 128$. And the weighting parameter $\zeta$ is fixed to $10^{-3}$. For parameters in Actor, the number of LSTM cell units is set to 256 and batch size is 64. In Critic, the number of hidden layer is 4. The batch size in replay buffer is 64 and we set 200 sessions in each episode. Besides, in training procedure, we set learning rate $\eta = 0.001$, discount factor $\rho = 0.9$, and the rate of target networks soft update $\tau=0.001$. 

\subsection{Comparisons and Analysis}
\newcommand{\tabincell}[2]{\begin{tabular}{@{}#1@{}}#2\end{tabular}}
\begin{table}
	\caption{Performance comparison on user identity linkage}
	\centering
	\begin{tabular}{l|c|ccccccc}
		\hline
		Dataset&Metrics&MAG&IONE&SiGMa&SDM&DeepLink&PAAE&RLink\\\hline
		\multirow{3}{*}{\tabincell{c}{Foursquare\\Twitter}}&\textit{P@1}&6.38&22.38&-&51.92&34.47&21.98&\textbf{75.93}\\\cline{2-9}
		&\textit{P@9}&17.05&46.38&-&51.92&66.09&47.62&\textbf{75.93}\\\cline{2-9}
		&\textit{MAP}&-&32.79&-&51.92&47.78&40.68&\textbf{75.93}\\\hline
		\multirow{4}{*}{\tabincell{c}{Last.fm\\Myspace}}&\textit{P@1}&-&29.57&\textbf{95.65}&72.72&-&28.76&85.71 \\\cline{2-9}
		&\textit{P@9}&-&53.21&\textbf{95.65}&72.72&-&52.97&85.71\\\cline{2-9}
		&\textit{MAP}&-&47.61&\textbf{95.65}&72.72&-&46.25&85.71 \\\cline{2-9}
		&\textit{recall}&-&-&32.86&6517&-&-&\textbf{77.68} \\\hline	
		\multirow{4}{*}{\tabincell{c}{Aminer\\Linkedin}}&\textit{P@1}&-&31.76&88.50&82.50&-&30.96&\textbf{92.85} \\\cline{2-9}
		&\textit{P@9}&-&59.28&88.50&82.50&-&59.42&\textbf{92.85} \\\cline{2-9}
		&\textit{MAP}&-&50.12&88.50&82.50&-&52.76&\textbf{92.85} \\\cline{2-9}	
		&\textit{recall}&-&-&47.39&79.46&-&-&\textbf{91.54} \\\hline	
	\end{tabular}
\label{performance}
\end{table}

We compare our proposed model RLink with the following recent user identity linkage methods. Note that SiGMa, SDM and RLink are prediction methods which determine whether two user identities from original and target are matching or not. Therefore, the evaluation of those methods becomes $\textit{P@k}=\textit{P@1}=\textit{MAP}$. From Table \ref{performance}, we can see that:
\begin{itemize}
	\item RLink is significantly better than previous user identity linkage methods. This result demonstrates that our method which considers UIL as a sequence decision problem and makes decisions from a global perspective is useful. Besides, reinforcement learning based methods, which learns to directly optimize the overall evaluation metrics, works better than other deep learning methods, such as DeepLink and PAAE. 
	\item SDM and SiGMa achieve higher precision than other baselines, which demonstrates that considering the influence of previous matched information is beneficial for UIL task. Although SigMa achieves the highest precision on Last.fm-Myspace, it suffers from a lowrecall due to its fixed greedy matching strategy. By contrast, SDM and RLink promotes about $30\%$ and $40\%$ recall over SigMa on respectively, which proves previous matched result has a long-time impact on the following matching process.  
	\item IONE, PAAE and DeepLink perform better than MAH and MAG, which means that considering more structural information, such as followership/ followee-ship and global network structure, could achieve higher precision. Besides, the precision of PAAE and DeepLink are higher than IONE, which demonstrates that considering global network structure and utilizing deep learning model could improve the performance of UIL.
	\item Besides, we note that all methods perform better on Aminer-Linkedin than they do on Last.fm-Myspace, maybe because the scale of the training data in Aminer-Linkedin is larger.     
\end{itemize}
\subsection{Discussion }
\subsubsection{Parameter Sensitivity}
In this section, we analyze the sensitivity of three parameters which are different $K$ in \textit{P@K}, training ratio $r$ and embedding dimension $d$. The performance of those parameters on all three testing datasets is similar. We present the performance on Twitter-Foursquare due to the limited spaces.  
\begin{figure*}[h]
	\centering
	\subfigure[different K]{
		\includegraphics[width=3.6cm ]{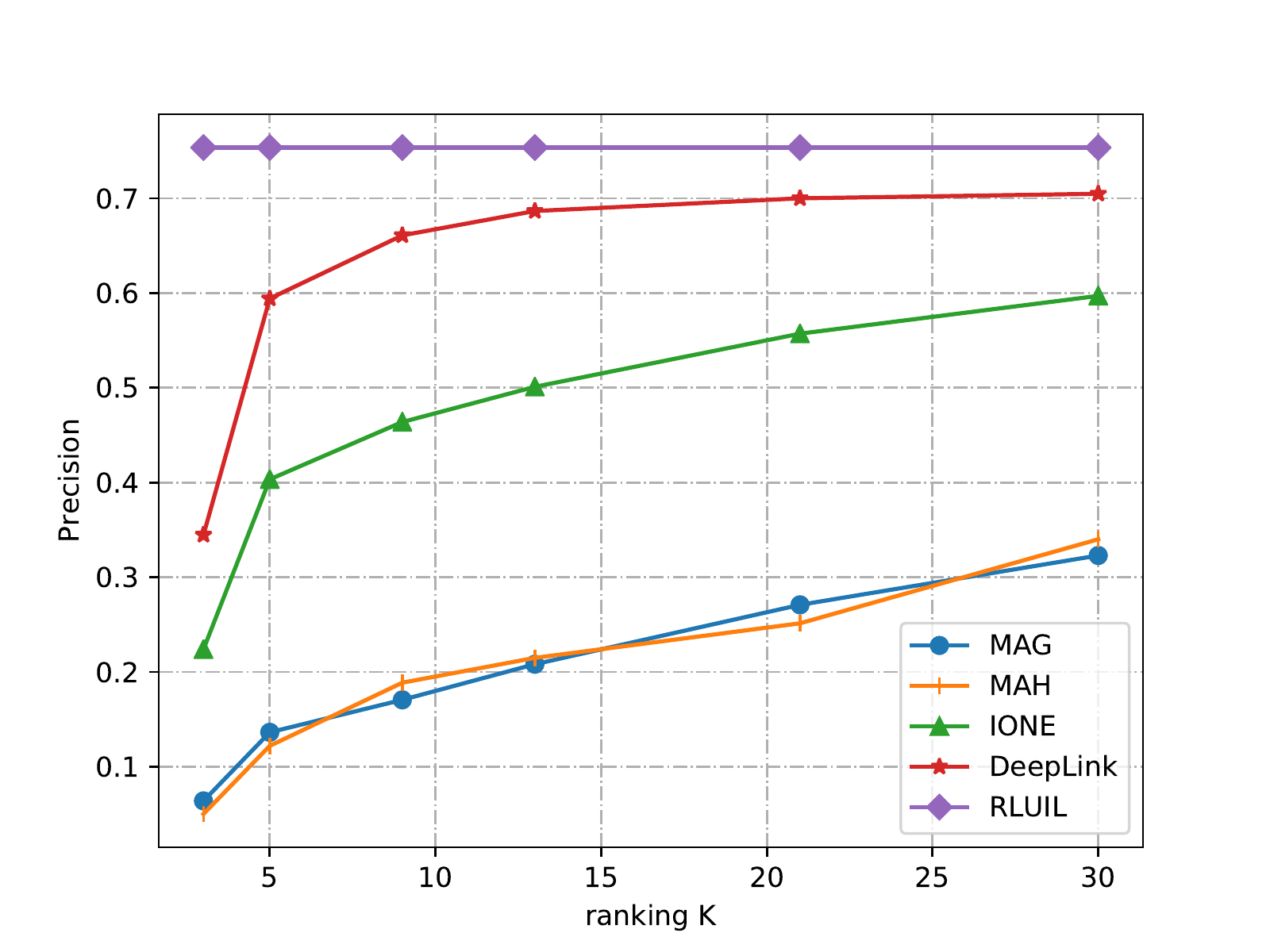}
	}
	\subfigure[Training Ratio]{
		\includegraphics[width= 3.6cm ]{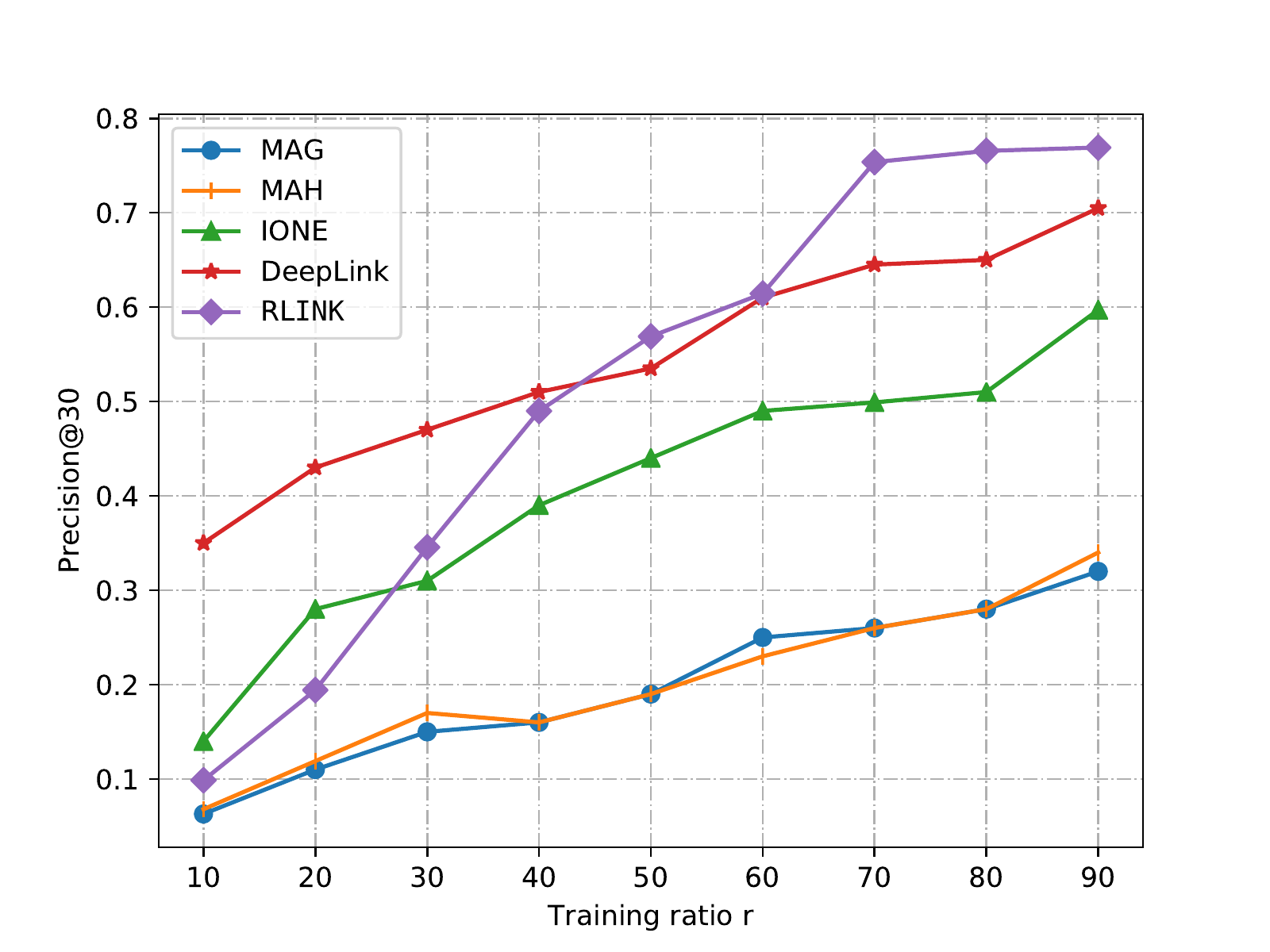}
	}
	\subfigure[Embedding Dimension]{
		\includegraphics[width= 3.6cm ]{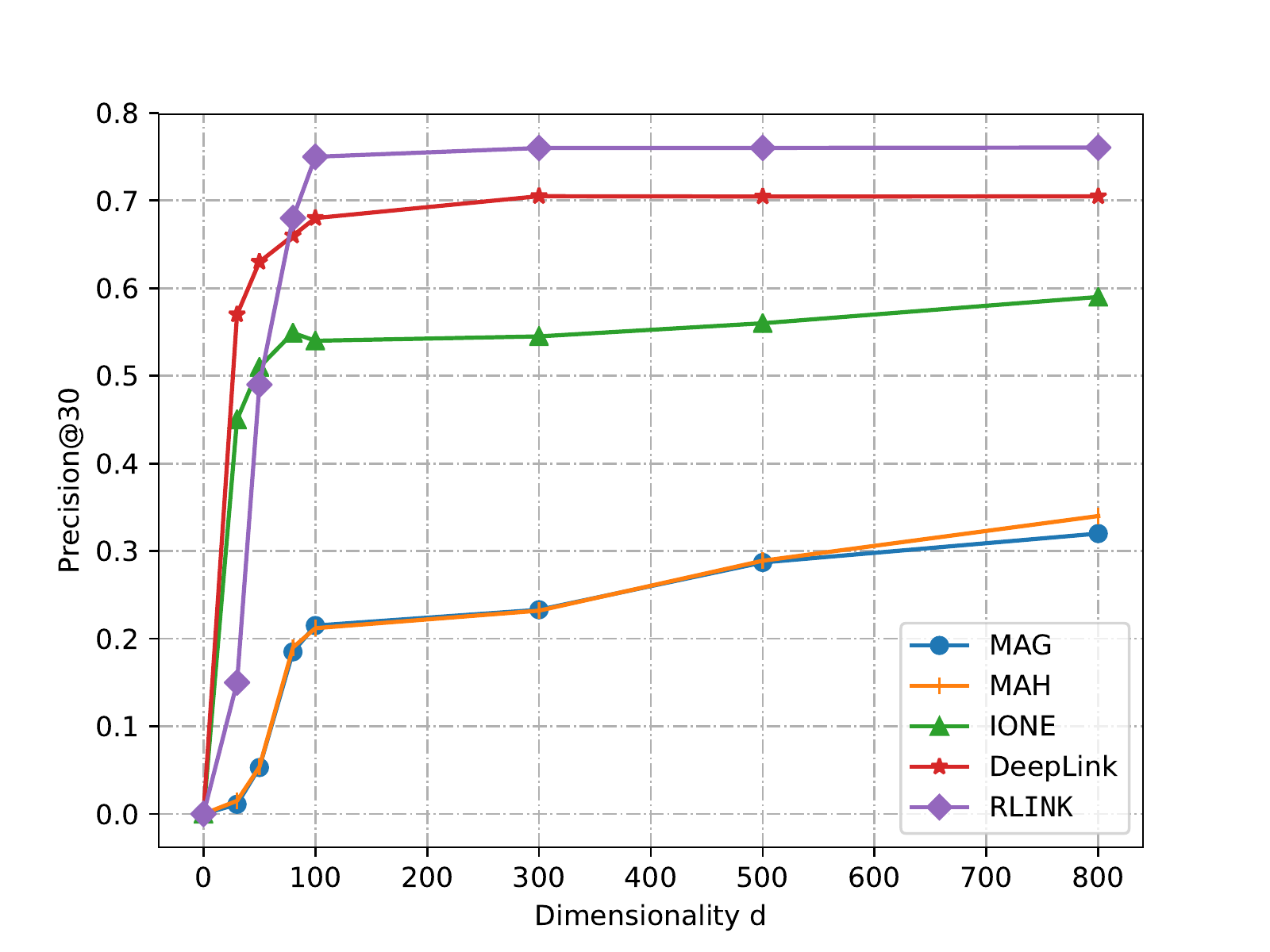}
	}
	\caption{Detailed Performance Comparison on Twitter-Foursquare Dataset}
	\label{four}
\end{figure*}

\textbf{Precision on different $K$.}
For different $K$ in \textit{P@K}, we reports the precision of different methods on variable $K$ between $1$ and $30$. RLink outperforms all the comparison methods consistently and significantly given different @K settings. IONE performs better than MAG and MAH, showing that constructing the incidence matrices of the hypergraph could fail to differentiate the follower-ship and followee-ship. Both RLink and DeepLink perform better than MAG, MAH and IONE, showing that deep learning methods could extract more node feature for UIL than traditional methods. The precision of most of the testing methods grow increase significantly with the increase of $K$ until $K=20$, indicating that most of ranking methods achieve matched identity in $Top-20$.

\textbf{Precision on different training ratio $r$.}
For different training ratios, we reports the Precision@30 of different methods on variable training ratio between $10\%$ and $90\%$.
RLink outperforms all the comparison methods when the ratio rose to $60\%$. The ratio of anchor nodes used for training greatly affects the performance of RLink. Especially for ratio settings as $60\%$ to $70\%$, the performance enhancement is significant. While the result shows that with the increase of training data, the precision of UIL firstly increases significantly and then does not increase drastically as the ratio rose to $70\%$. And the comparative methods achieve good performance when the training ratio is setting around $90\%$, which demonstrates the robust adaptation of our method RLink. Besides, from Figure \ref{four}b, we found that DeepLink performs better than 
RLink when the training ratio is less then $40\%$, indicating that RLink might need more train dataset than DeepLink. 

\textbf{Precision on different embedding dimension $d$.}
According to Figure \ref{four}c, both MAG and MAH achieve good performance when the dimensionality setting is around 800, while RLink, IONE and DeepLink achieve good performance when the dimensionality is around 100. The complexity of the learning algorithm is highly depending on the dimensionality of the subspace. And low dimensional representation also leads to an efficient relevance computation \cite{IONE}. Therefore, we conclude that RLink, IONE and DeepLink is significantly more effective and efficient than MAH and MAG. Besides, RLink achieves the best results when the dimensionality setting is bigger than 80.

\subsubsection {Evaluation of Actor-Critic Framework}
\begin{figure}[htbp]
	\centering
	\subfigure[Last.fm-Myspace]{
		\includegraphics[width=5cm]{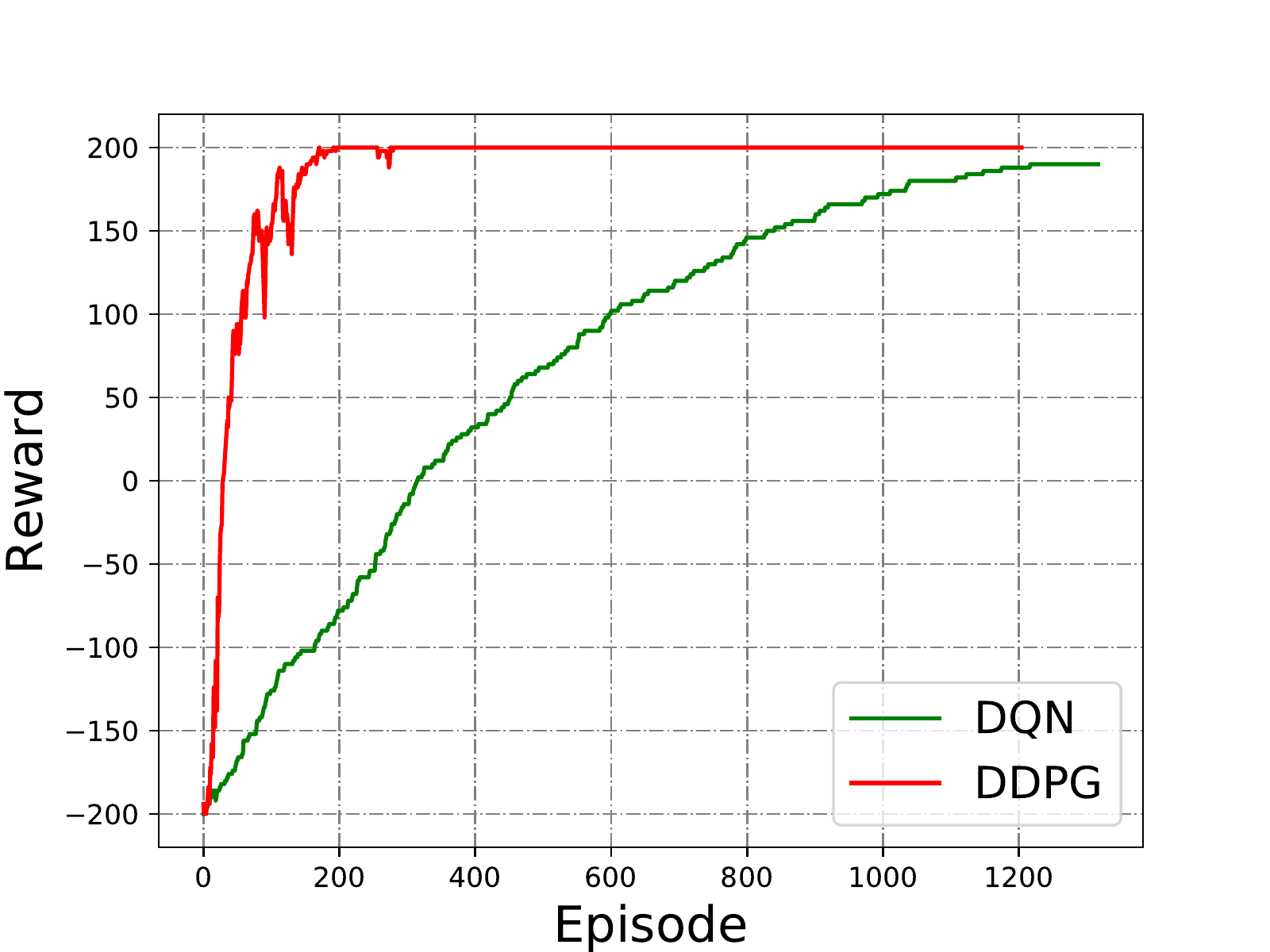}
	}
	\subfigure[Linkedin-Aminer]{
		\includegraphics[width=5cm]{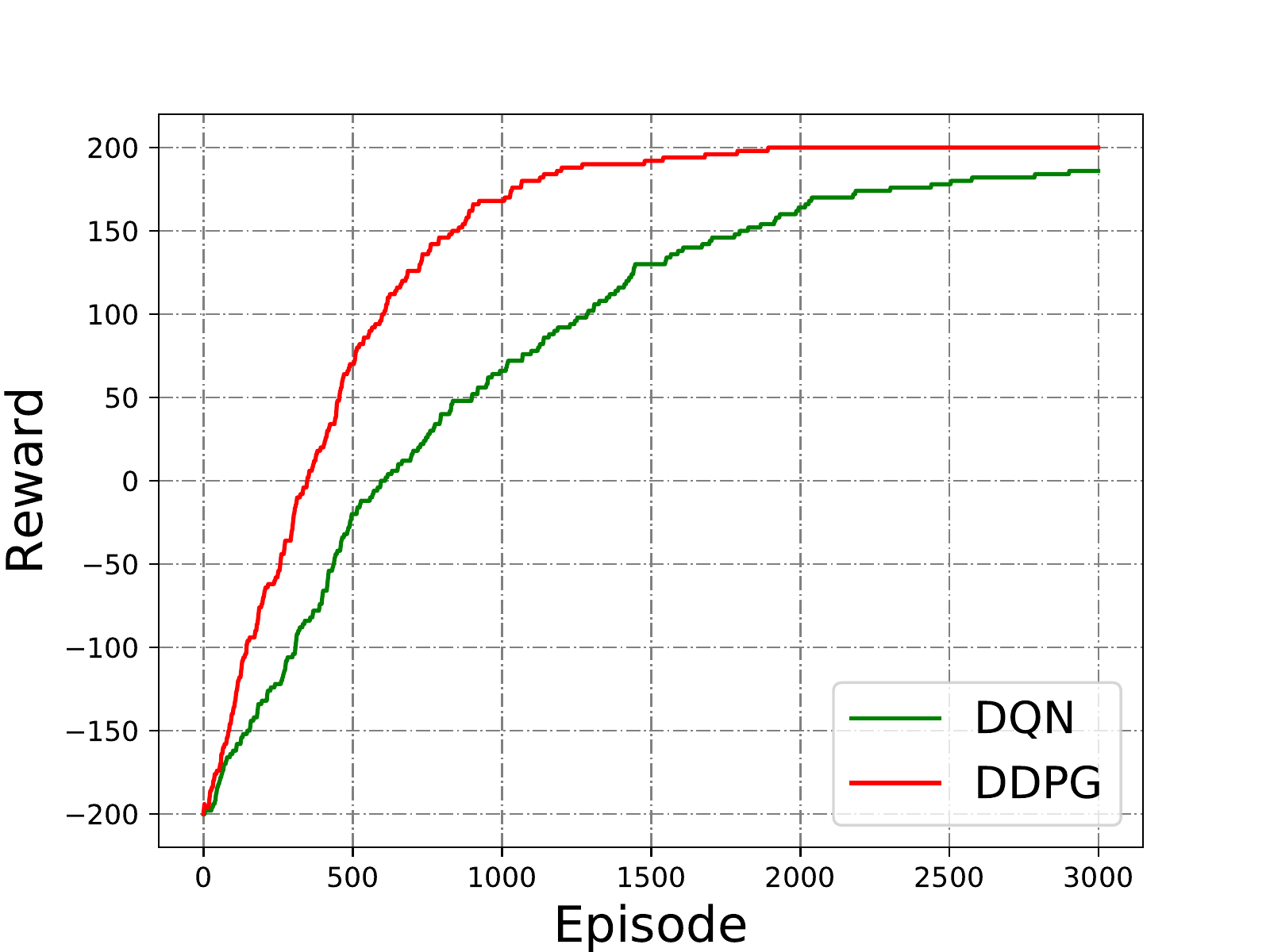}
	}
	\caption{Comparison between DDPG and DQN: The x-axis shows the training episodes. The y-axis shows the total reward of each episode. Red line represents DDPG and green Line is DQN.}
	\label{reward}
\end{figure}

In this paper, we use DDPG algorithm to train the RLink model. To evaluate the effectiveness of the Actor-Critic framework, we compare the performance of DDPG with DQN. From Figure \ref{reward}, we can see that DQN performs similar to DDPG, but the training speed of DQN is much slower. As shown in Figure\ref{reward} (a), DQN needs 1200 episodes to converge, which is almost four times of the episodes DDPG needs. In Linked-Aminer, as shown in Figure\ref{reward}(b), both DDPG and DQN need more training episodes, i.e., 750 and 2000 episodes respectively, when the size of action space rose to $4153\times 4153$. Besides, DQN performs worse than it does in Last.fm-Myspace, which demonstrates that DQN is not suitable for the large action space. In summary, DDPG achieves better performance and faster training speed than DQN, which indicates that Actor-Framework is suitable for UIL with enormous action space. 

\subsubsection{Effect of Long-term Reward}
\begin{figure}[htbp]
	\centering
	\subfigure[Last.fm-Myspace]{
		\includegraphics[width=5cm]{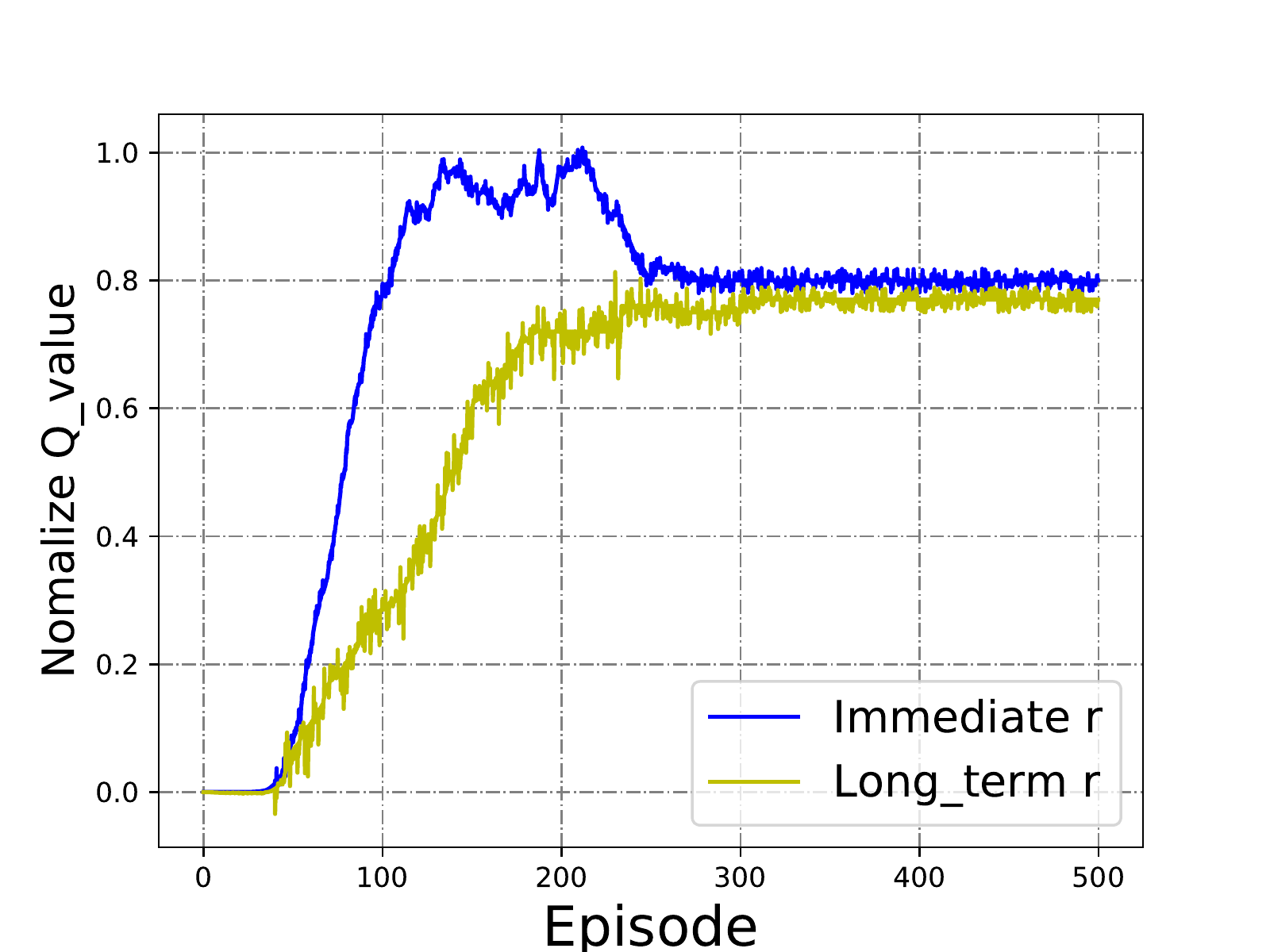}
	}
	\subfigure[Linkedin-Aminer]{
		\includegraphics[width=5cm]{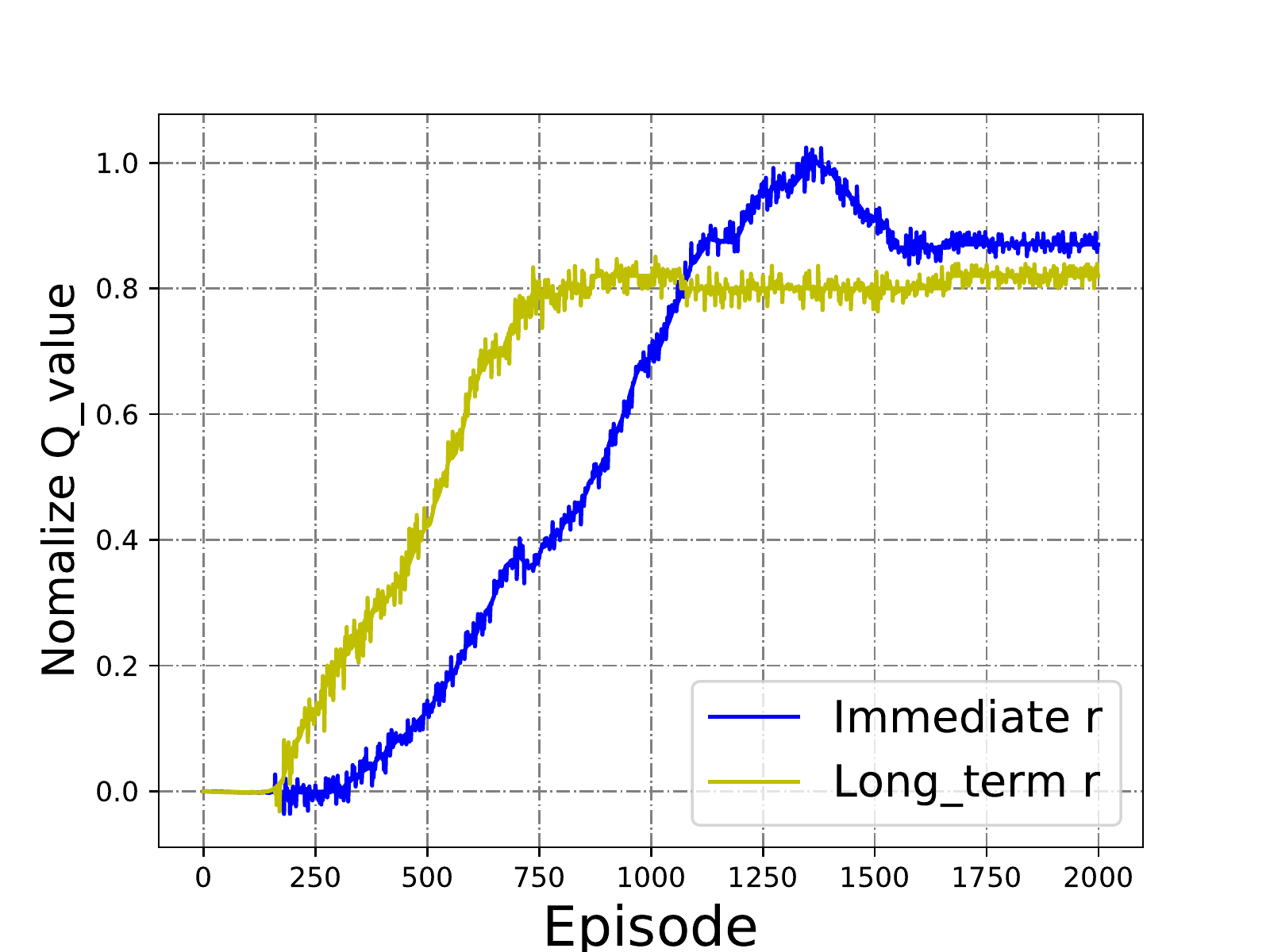}
	}
	\caption{Q-value performance with different reward on Last.fm-Myspace and Linkedin-Arminer. The x-axis shows the training episodes. The y-axis shows the normalize Q-value of each session. The value of immediate reward is equal to $1/-1$, while long-term reward is $\frac{1}{t}/\frac{-1}{t}$ }
	\label{iteration}
\end{figure}

To evaluate the effectiveness of long-term reward, we compare it with an immediate reward via Q-value performance. Q-value performance is a judgment of whether $a$ suits $s$. From Figure \ref{iteration}, we can find that both long-term and immediate reward would converge to a similar value, but the Q-value of immediate reward is increased first and then decreased. The higher Q-value in the training procedure indicates that the immediate reward may get trapped in a local optimum. By contrast, the long-term reward makes our model make decisions from a global perspective and the Q-value smoothly converge.

\subsubsection{Effectiveness of RLink Components}
This experiment is designed to validate the effectiveness of main components in the Actor network, including the input features, attention layer, LSTM layer, action transformer and decoder. We systematically eliminate the corresponding component and define the following variants of RLink.
\begin{itemize}
	\item $RLink_{LINE}$ (RLink with LINE as pre-trained embedding method): In this variant, we replace the pre-trained embedding method Node2vec by LINE\cite{line} to evaluate the diversity of different pre-trained identity embeddings.
	\item RLink-FEED (RLink without feedback information): This variant is to evaluate the feedback fed into the Encoder. So, we just use the representation of matched identity pair as the history matching information.
	\item RLink-LSTM (RLink without LSTM layer): This variant is to evaluate the contribution of LSTM layer. We replace the LSTM layer by simple fully-connected layer.
	\item RLink-ATT (RLink without attention machine): This variant is to evaluate the contribution of the attention layer. So, we remove the attention machine after the LSTM layer. 
	\item RLink-TRANS (RLink without transformer layer): In this variant, we replace the action transformer component by fully-connected layer to evaluate its effectiveness. 
	\item $RLink_{MLP}$ (RLink utilizes MLP as decoder): In this variant, we replace the single neural network (NN) layer  in the decoder by MLP to evaluate the effective of NN layer.
\end{itemize}

\begin{table}
	\caption{Performance on different RLink components}
	\begin{tabular}{ccccl}
		\toprule
		Datasets & \small{Methods}& \small{Precision} & \small{Recall} & \small{F1} \\
		\midrule
		&\small{$RLink_{LINE}$} & 0.8511& 0.7732& 0.8014\\    
		&\small{RLink-FEED} &0.7055& 0.7182& 0.7118  \\
		Last.fm-&\small{RLink-LSTM} & 0.8014& 0.7189& 0.7579 \\
		Myspace&\small{RLink-ATT} & 0.7921& 0.7312&0.7604 \\
		&\small{RLink-TRANS} & 0.7155& 0.6367& 0.6738 \\
		&\small{$RLink_{MLP}$} & 0.8297& 0.7354& 0.7797\\
		&\small{RLink} & \textbf{0.8571}& \textbf{0.7768}& \textbf{0.8038} \\
		\bottomrule
		&\small{$RLink_{LINE}$} & 0.9213& 0.8987& 0.9099 \\   
		&\small{RLink-FEED} & 0.8372& 0.8265& 0.8319 \\
		Aminer-&\small{RLink-LSTM} & 0.8648&0.8434& 0.8540 \\
		Linkedin&\small{RLink-ATT} & 0.8726& 0.8573& 0.8649 \\
		&\small{RLink-TRANS} & 0.8439& 0.8191& 0.8313 \\
		&\small{$RLink_{MLP}$} & 0.9076& 0.8879& 0.8976 \\    
		&\small{RLink} & \textbf{0.9285}& \textbf{0.9027}& \textbf{0.9154} \\
		\bottomrule
	\end{tabular}
	\label{components}
\end{table}

The result is shown in Table \ref{components}. We can find that    $RLoml_{LINE}$ adapt LINE to pre-train user identity embedding vectors, which achieves similar performance to RLink. This phenomenon indicates that the pre-trained identity embeddings could not have great influence on the performance of UIL. RLink-FEED performs the worst among all variants, which demonstrates that the feedback information significantly promotes the performance of our model. RLink-LSTM performs worse than RLink, which suggests that capturing the long-term memory of the history matched information by LSTM is very beneficial for the subsequent matching. RLink-ATT proves that incorporating attention mechanism can better capture the influence of each previous decision than only LSTM. RLink-TRANS proves that action transformer could map $a^{val}$ and $a^{cur}$ exactly. $RLink_{MLP}$ proves that simply neural network in the decoder phase could perform better than more complicated model (MLP). 
RLink outperforms all its variants, which indicates the effectiveness of each component for UIL.         

\section{Related Work}

In this section, we briefly introduce previous researches related to our study, including user identities linkage and reinforcement learning. 
\subsection{User Identities Linkage}
Previous UIL works consider UIL as a matching problem, or utilize classification models and label propagation algorithms to tackle this task. Matching-based methods build a bipartite graph according to affinity score of each candidate pair of identities and achieves one-to-one matching for all user identity pairs based on this bipartite graph \cite{53,34}. The basic principle is Stable Marriage Matching. Classification-based models classify whether each candidate matching is correct or not. Commonly used classifiers include Naive Bayes, Decision tree, Logistic regression, KNN, SVM and Probabilistic classifier\cite{23,51,25,PALE}. Label-Propagation based methods discover unknown user identity pairs in an iterative way from the seed identity pairs which have been matched\cite{69,32,48,40,78}. Some recent works prefer to combine above methods, for example, \cite{cosnet} computes local consistency based on matching model and applies label propagation for global consistency. 

With the development of network embedding and deep learning, embedding based methods and deep learning models have been utilized to solve UIL problem. Embedding based methods embed each identity into the low-dimensional space which preserve the structure of network firstly, and then align them via comparing the similarity between embedding vectors across networks \cite{PALE,MAH,deeplink,Ulink}. However, those two-step methods needs two subject which is difficult to optimize. \cite{IONE} proposes an unified framework to address this challenge, where the embeddings of multiple networks are learned simultaneously subject to hard and soft constraints on common users of the network. 
\cite{active} introduces an active learning method to over the sparsity of labeled data, which utilizes the numerous unlabeled anchor links in model building.
Finally, inspired by the recent successes of deep learning in different tasks, especially in automatic  feature extraction and representation, \cite{deeplink} propose a deep neural network based algorithm for UIL.     

\subsection{Reinforcement Learning}
Reinforcement Learning (RL) is one of the most important machine learning methods, which gets optimal policy through trail-and-error and interaction with dynamic environment. Generally, RL contains two categories: model-based and model-free. And the most frequently used method is model-free reinforcement learning methods which can be divided into three categories: Policy-based RL, Q-learning and Actor-Critic. The policy-based RL \cite{fangzheng,policy} learns the policy directly which compute probability distribution of every action.
The Q-learning \cite{Atari,hybrid} learns the value function which evaluate the value of each potential action-state pair. And Actor-Critic \cite{Deeppage,actor-critic} learns the policy and value function simultaneously which aims to evaluate the quality of a deterministic action at each step. 

Recently, due to various advantages of RL, it has been successfully applied in many fields, such as Game \cite{Atari,Go}, Computer Vision \cite{CV} and Natural Language Processing \cite{fangzheng}. However, due to the complexity of online social network analysis tasks, works based on RL is less than in other fields. \cite{dynamic} is an early work based on reinforcement learning in social network which argues modeling network structure as dynamic increases realism without rendering the problem of analysis intractable. Existing methods utilize DQN or Q-learning framework, such as \cite{graph} applies DQN to address Graph Pattern Matching problem and \cite{Peyravi} applies Q-Learning to search expert in social network. Inspired by the above works, we consider UIL as a markov decision problem and apply the reinforcement learning framework. However, computing Q-value of each identity pair is time-consuming. So, we adapt Actor-Critic framework to address UIL in this paper.     

\section{Conclusion} 
In this paper, we consider user identity linkage as a sequence decision problem and present a reinforcement learning based model. Our model directly generate a deterministic action based on previous matched information via Actor-Critic framework. By utilizing the information of previously matched identities and the social network structures, we can optimize the linkage strategy from the global perspective. In experiments, we evaluate our method on Foursquare-Twitter, last.fm-Mysapce and Linkedin-Aminer datasets, the results show that our system out performs state-of-the-art solutions. There are many other information in the network, such as user attribute information, and the links in the network is varied, such as AP (Author-Paper) and PC (Paper-Conference). Therefore, in the future, we would utilizes those attribute information and varied links in our model to address the UIL task.

%
%


%
%



\end{document}